# Identifying the differences between 3 dimensional shapes Using a Custom-built Smart Glove

Davis Le, Sairam Tangirala, and Tae Song Lee
Georgia Gwinnett College,
1000 University Ctr ln, Lawrenceville,
GA 30043

**Abstract**
Sensor-embedded glove systems have been reported to require careful, time-consuming, and precise calibrations on a per-user basis in order to obtain consistent, usable data[1]. We have developed a low-cost, flex-sensor based smart glove system that may be resilient to the common limitations of data gloves. This system utilizes an Arduino-based microcontroller as well as a single flex sensor on each finger. Feedback from the Arduino's analog to digital converter can be used to infer object's dimensional properties; the reactions of each individual finger will differ with respect to the size and/or shape of a grasped object[1,2,3]. In this work, we report its use in statistically differentiating stationary objects of spherical and cylindrical shapes of varying radii regardless of the variations introduced by glove's users[1]. Using our sensor-embedded glove system, we explored the practicability of object classification based on the tactile sensor responses from each finger of the smart glove. An estimated standard error of the mean was calculated from each of the of five fingers' averaged flex sensor readings. Consistent with the literature[2,3], we found that there is a systematic dependence between an object's shape, dimension and the flex sensor readings. The sensor output from at least one finger, indicated a non-overlapping confidence interval when comparing spherical and cylindrical objects of the same radius. When sensing spheres and cylinders of varying sizes, all five fingers had a categorically varying reaction to each shape. We believe that our findings could be used in machine learning models for real-time object identification.

**Introduction**
Physical objects have unique names, properties, and other qualities associated with them. Generally, humans can assess an object's qualities quickly using their many senses. However, the application of these processes to usable machines and devices has been an active area of research and development due to the sheer complexity of the task.[1] One method of translating these results is by defining the object via the implementation of the human capabilities of logical reasoning and sensing in these devices through object classification. Object classification is generally associated with computer vision and image processing techniques due to the simplicity of merely gazing at an object for the inference of its properties.[2] However, the human hand is both the counterpart of and the complement to the human eye. Where the eyes cannot see, the hands can "sense"; it is one of the major recipients of the human sense of touch.[3] When a human hand grasps an object, multiple highly complex kinematic, dynamic, and sensory interactions occur between the hand and an object.[3,5]

In low-visibility environments, contact-based, tactile sensing can supplement or outright replace vision-oriented object classification techniques.[4] The sense of touch could be emulated by tactile sensors, providing information that is unattainable by vision alone.[1] Glove systems with embedded tactile sensors are collectively considered data gloves;[6] these gloves generally feature multiple sensors on each finger to record and convey more information than a single sensor can by itself. As seen in **Figure 1,** one of the types of tactile sensors used in these data gloves are flex sensors. Flex sensors are variable resistors that provide a linear increase in electrical resistance as the flex sensor increases in bend angle[7]. In practice, when it is in a closed circuit, the flex sensor will change a voltmeter or ohmmeter's measurements when the flex sensor is bent. This is utilized to translate physical information from a finger's bend angle to digital information using an analog-to-digital converter on a microcontroller such as an Arduino.[6-9,11]

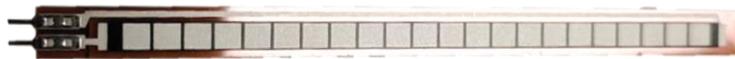

**Figure 1:** Image of a 112.5mm Spectra Symbol resistive flex sensor.[7]

Currently, data gloves are used for modelling human grasping behaviors,[5,6] biomedical or psychological research,[10] and human-machine interfaces for controlling various devices or interacting with virtual realities.[6,8] In regards to object-glove interaction, much of the focus is on observing or classifying dynamic grasp behaviors as a function of time.[5,6,11] Many of the different glove systems used in the literature have common limitations; they vary in number of sensor, sensor placements in the glove and require time-extensive calibration, in addition to varying performances due to differences in the hand geometry of differing users.[6,11] The sensor's performance depends on different users' hands and sensor's data varies in voltage output and the bend angle of all fingers of the data glove, causing grasp recognition rates to drop when observing multiple users as a whole.[6] Additionally, other research groups have reported object properties through flex sensors in the past, mostly in the context of robotics,[4,8] or limiting the object classification to the use of three fingers within a data glove.[11]

In this work, we demonstrate the glove's ability to differentiate objects of varying sizes and shapes using a data glove with a single flex sensor embedded in all five fingers; we observe static, non-motile object grasping to explore the feasibility of object classification regardless of users' hand differences. Analyzing the data glove's behavior during contact with the object eliminates variations in flex sensor readings as a function of time and flex sensor movement.[8,11] The use of all five fingers provides a larger dataset when observing object properties, as individual fingers will differ in flex sensor readings when interacting with differing shapes[11], as opposed to robot-mounted flex sensor object readings, which yield similar readings across sensors due to the design of their moving parts.[4,8] However, our work includes rescaling collected data between differing users through min-max normalization for establishing an averaged dataset. The collated averages undergo standard error analysis to establish clear differences between object sizes and shapes, accounting for and potentially resolving issues caused by data glove use with multiple users as seen in previous works.[6]

## Methods

*Smart Glove System*

We chose to use an Arduino UNO microcontroller (**Figure 2**) to digitize the flex sensor output. In this case, it is used to interface with analog and digital hardware, using Arduino Software or IDE. The 10-bit analog-to-digital converter (ADC) and the internal voltmeter functions of the Arduino converter allow for the conversion of a 0V-5V (voltage) input to integer values of 0-1023. The one unit of digitized output corresponds to approximately 4.9 mV. Additionally, this platform can convert an analog input to a digital output in one millisecond, which was used in this experiment to mitigate sources of error.[12] The Arduino shown in **Figure 2A** is connected to five identical flex sensors (**Figure 1**). A resistor is included in circuit to modulate ADC value outputs.

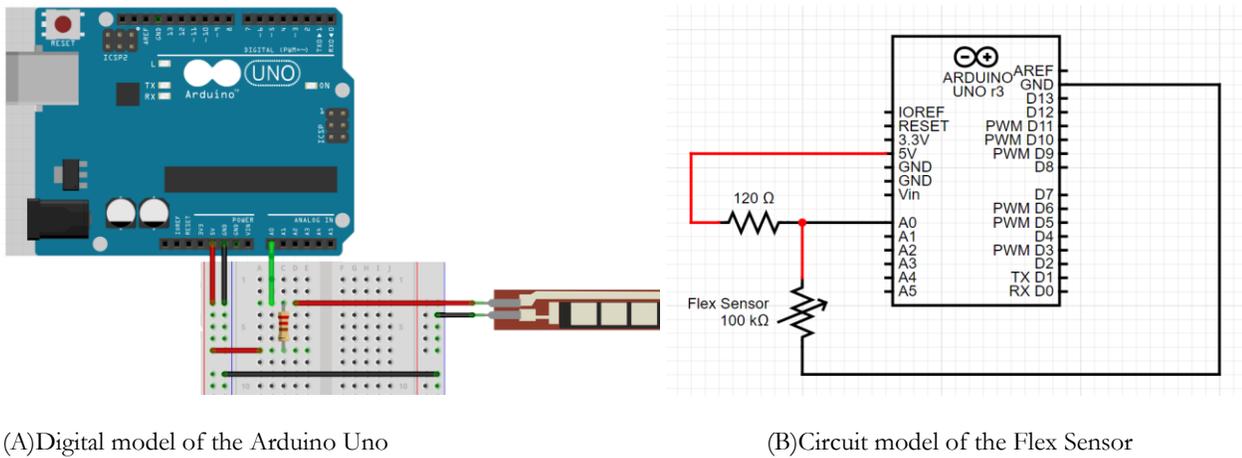

(A) Digital model of the Arduino Uno  (B) Circuit model of the Flex Sensor

**Figure 2 A,B**: Two digital representations of the wiring diagram for each flex sensor. These diagrams were generated using Fritzing and Tinkercad.[13,14]

The five flex sensors used in **Figure 3** are connected to Analog Input pins A4, A0, A1, A2, and A3 for the Thumb, Index, Middle, Ring, and Pinky finger respectively; these sensors are mounted onto each of a commercial glove's fingers. Each flex sensor has wiring nearly identical to the **Figure 2** digital models, with the only change being the pin connections to the Arduino. This system is configured for collecting ADC values corresponding to resistance values, which are sent to an external computing device.

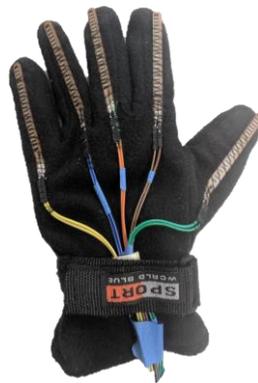

**Figure 3**: Image of a custom-built data glove with five separate, identical 112.5mm flex sensors mounted on each finger.

*Data Collection*

ADC values were separated for observation on an individual, single-finger basis in order to analyze the data glove's individual finger behavior with a corresponding object. When the smart glove was in use, all finger-mounted flex sensors' ADC values were collected simultaneously in 50 millisecond intervals for five seconds, resulting in 100 data points per finger; the data collected was stored in Microsoft Excel files for averaging and data analysis. These per-finger ADC values were averaged to account for sources of error such as user movement; they are treated as single ADC values associated with each user.

*Single Flex Sensor characterization using a semi-rigid polymer ring*

The behavior of a single flex sensor was observed for the determination of its performance and reliability at multiple, fixed bending diameters. The flex sensor, with a circuit as described in the *Smart Glove System* subsection, was mounted onto an adjustable, semi-rigid polyethylene ring (**Figure 3A**).

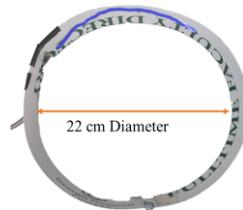

(A) Adjustable Polyethylene Ring

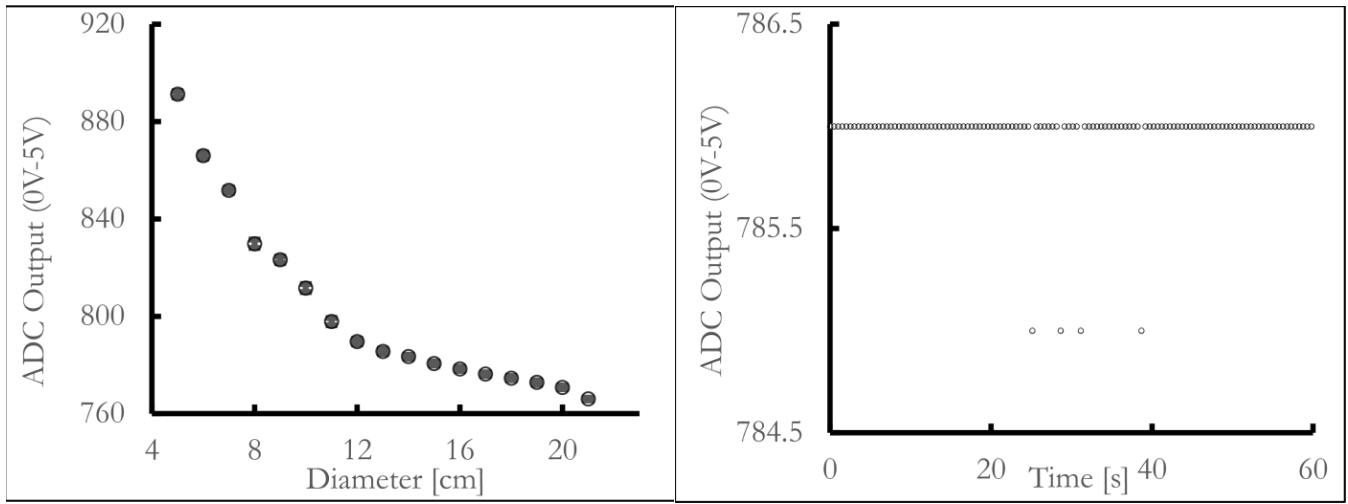

(B) Single Flex Sensor on Adjustable Ring    (C) Single Flex Sensor at a Fixed Diameter of 12 cm

**Figure 3**: Single Flex Sensor Characterization Using a Semi-Rigid Polymer Ring

On the left, in **Figure 3B**, the ring assembly was used to collect data over 5 trials, which is shown in the graph as averages per plot point. This ring was reduced in in size by 1-centimeter increments, starting at 22 cm and ending at 5 cm; a data point was collected and generated for every 1 cm increment. Five trials were performed in order to minimize sources of error. Error bars using standard error calculations are also shown in this figure. The behavior of the flex sensor changes at 12 cm in ring diameter. With a greater increase in ring diameter, flex sensor bend decreases and the corresponding voltage decreases.

**Figure 3C**, on the right, depicts a plot showing the time dependence of an individual flex-sensor's resistance for a fixed radius of curvature. Here, 1,000 data points were collected for over 8 minutes to study the flex sensor's output variation as a function of time, limited to ±1 fluctuations in ADC output.

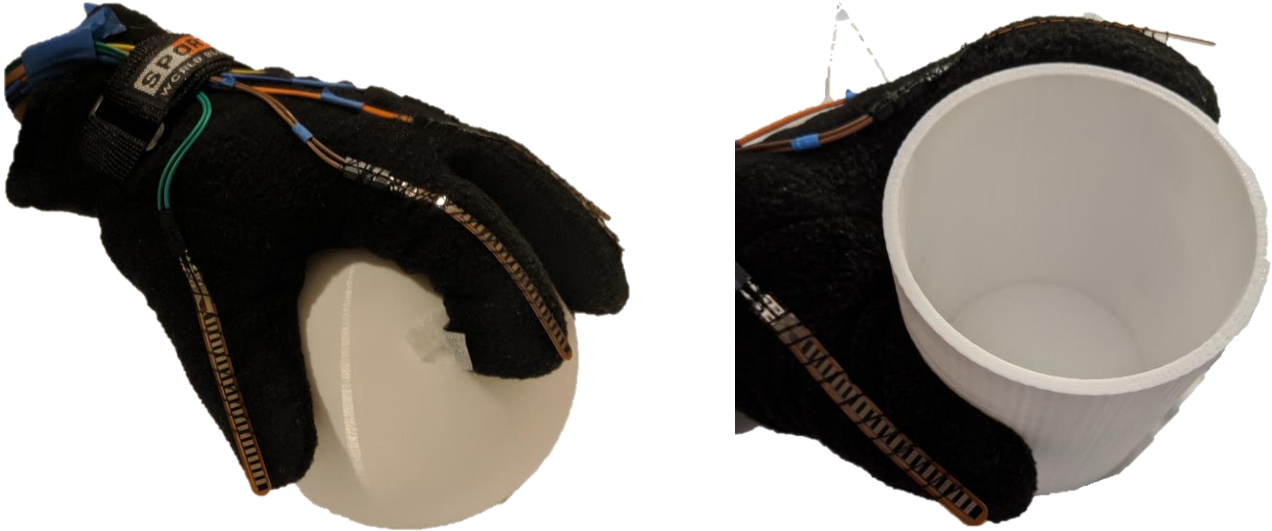

(A) Glove-Sphere interaction                (B) Glove-Cylinder interaction

**Figure 5:** Images of object-data glove grasping processes. Each of the 21 objects were 3D printed from polylactic acid filament. The spheres and cylinders, respectively, are identical in all parameters except diameter.

*Glove-Object Interaction*

To maintain consistency with standard grasping methodologies[6], all users performed only spherical and cylindrical grasping types. Several spheres and cylinders were 3D printed and their radius varied in 1-centimeter increments to observe the effects of object size on flex sensor behavior. For spherical objects, eleven separate users were asked to place the worn data glove on nine different spheres ranging from 6 cm to 16 cm diameter. When the fingertips and palm of the glove were in contact with the sphere as seen in **Figure 5A**, processes of the *Data Collection* subsection were performed. When testing cylinders, eight separate users were asked to place the data glove on ten different cylinders ranging from 6 cm to 16 cm diameter. In **Figure 5B**, the cylinders were treated the same as the spheres; *Data Collection* processes occurred after the fingertips and palm of the glove contacted the object.

*Data Analysis*

Each raw ADC value collected for a given shape and size underwent linear transformation using min-max normalization[15] to account for data glove calibration between differing users.[6] For a given user, each of their individual fingers' ADC values were normalized between the minimum and maximum diameters of the shape. The normalized values were then collated based on object size and shape. For example, all pinky finger values collected from interaction with an 8 cm cylinder were averaged together.

Afterwards, the collected data was plotted as a function of averaged, normalized ADC values over object size on a per-finger basis to observe individual finger behavior with a change in size. Plotted values were compared through statistical modelling techniques; linear regression was used to determine the relationship between ADC values and object diameter, whereas standard error of the mean (SEM) calculations were performed to observe confidence intervals between the data points.[15]

**Results**

*Smart Glove-Sphere Interaction*

The following results compare all five fingers from Smart Glove-Sphere interaction. The plot shown in **Figure 6** depicts averaged and linearly transformed ADC values collected from all fingers as a function of diameter.

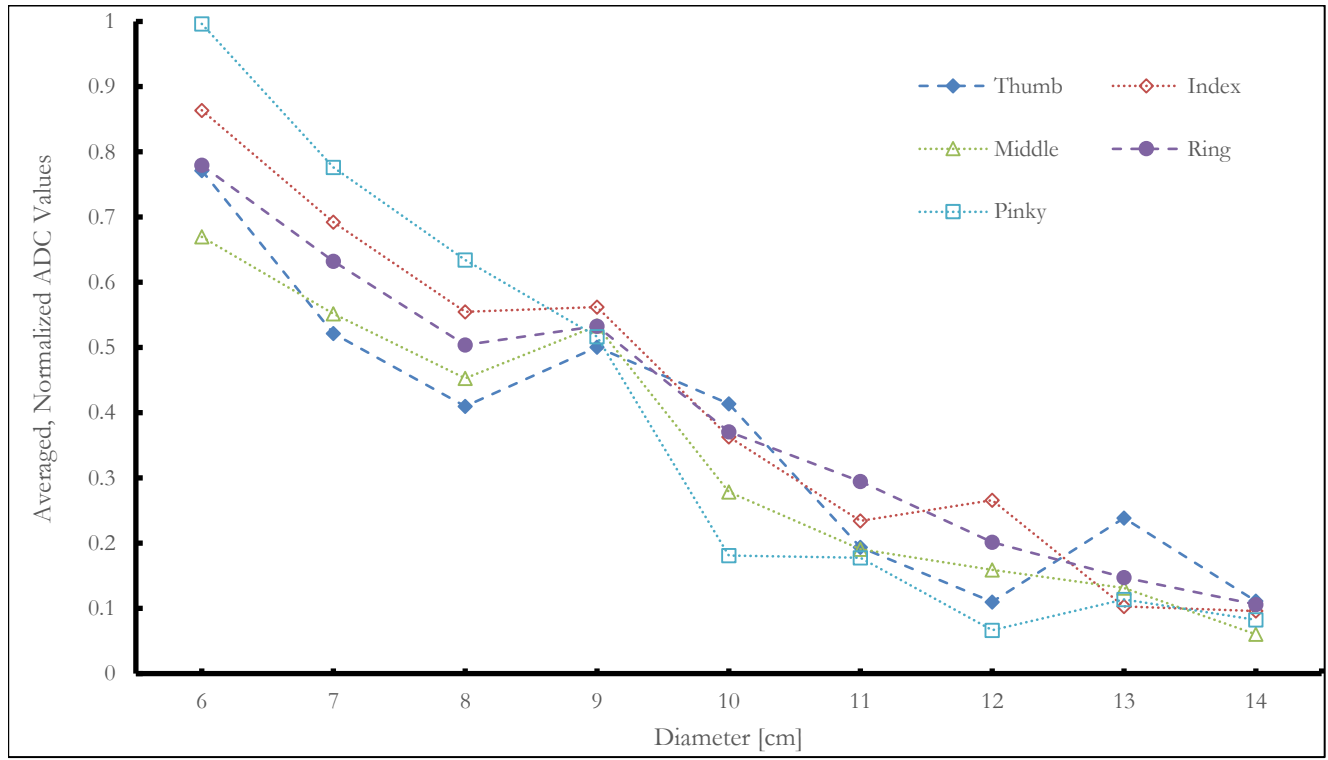

**Figure 6**: Smart Glove-Sphere Interaction

The flex sensors' behavior overall is akin to the observations found in the *Single Flex Sensor* subsection, where an increase in diameter decreases the read ADC values in correlation. However, each finger has a unique reaction with the spheres as they decrease in size. No other finger except the Ring finger has an exact value of 1 at a sphere diameter of 6 cm; all 11 users' tested Pinky finger ADC values were at their maximum value at the smallest sphere. The Thumb, Middle, Ring, and Index fingers increase in value at 9 cm Sphere diameter. Lastly, the Middle finger has the smallest change in digitized output between Sphere sizes.

*Smart Glove-Cylinder Interaction*

This subsection describes ADC values collected for the Cylinder as a function of diameter, seen in **Figure 7**. A 10 cm cylinder was not available for testing at the time of this writing; 10 cm data was not included in the provided plots. Like previous subsections *Single Flex Sensor* and *Smart Glove-Sphere Interaction*, the flex sensors for all fingers decrease in ADC value as the object diameter increases.

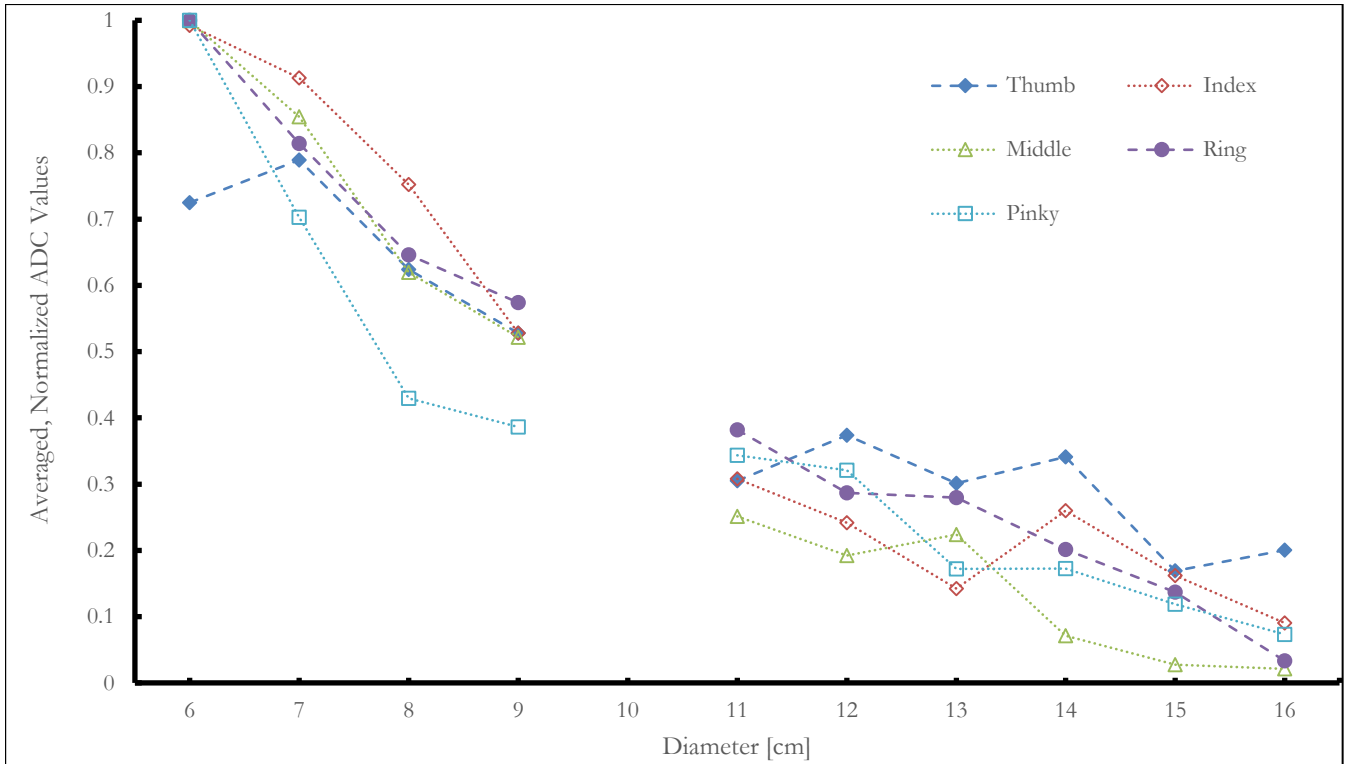

**Figure 7**: Smart Glove-Cylinder Interaction

Unique to Smart Glove-Cylinder Interaction, the Index, Middle, Ring, and Pinky fingers all start at their maximum values. Even with 8 different users, all fingers (except Thumb) for all tested users generally had the same reaction to the smallest cylinder. Within Cylinders larger than 11 cm, the Thumb has higher normalized ADC values than other fingers; it also changes the least across all diameters.

*Individual Finger Analysis*

The following subsection introduces five plots (**Figures 8-12**) comparing each individual fingers' ADC values between all Sphere and Cylinder sizes. Each averaged, normalized data point has included error bars based on SEM values generated from each dataset. There are more Cylinder diameters than Sphere diameters, but the 10 cm Cylinder is also missing from these plots as stated earlier. The sample sizes for each shape also differed; there were 11 users for collected Sphere data, and only 8 users for Cylinder data in comparison. The averaged, normalized ADC values from flex sensors on each finger had a differing reaction to each shape.

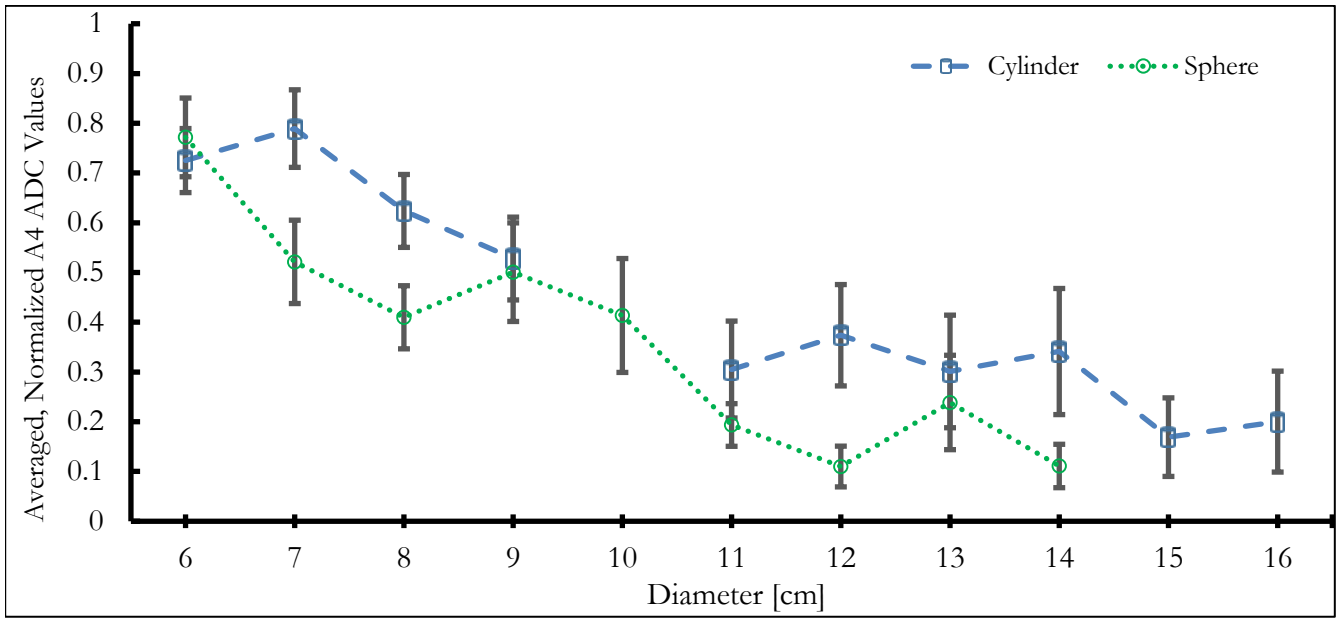

**Figure 8**: Thumb-Cylinder and Thumb-Sphere Interaction

**Figure 8** depicts change in Thumb flexion (decrease in finger bend angle) as a function of increasing object diameter. At 6 cm diameter, Cylinder ADC values were lower than Sphere ADC values. Conversely, at all diameters greater than 6 cm, all ADC values of Sphere were lower than Cylinder values. Overlap between error bars occur at 6, 9, and 13 cm. When observing 10-16 cm, the Thumb ADC values did not have a strong linear relationship with shape diameter for both shapes; $R^2$ values for objects greater than 10 cm were less than 0.50.

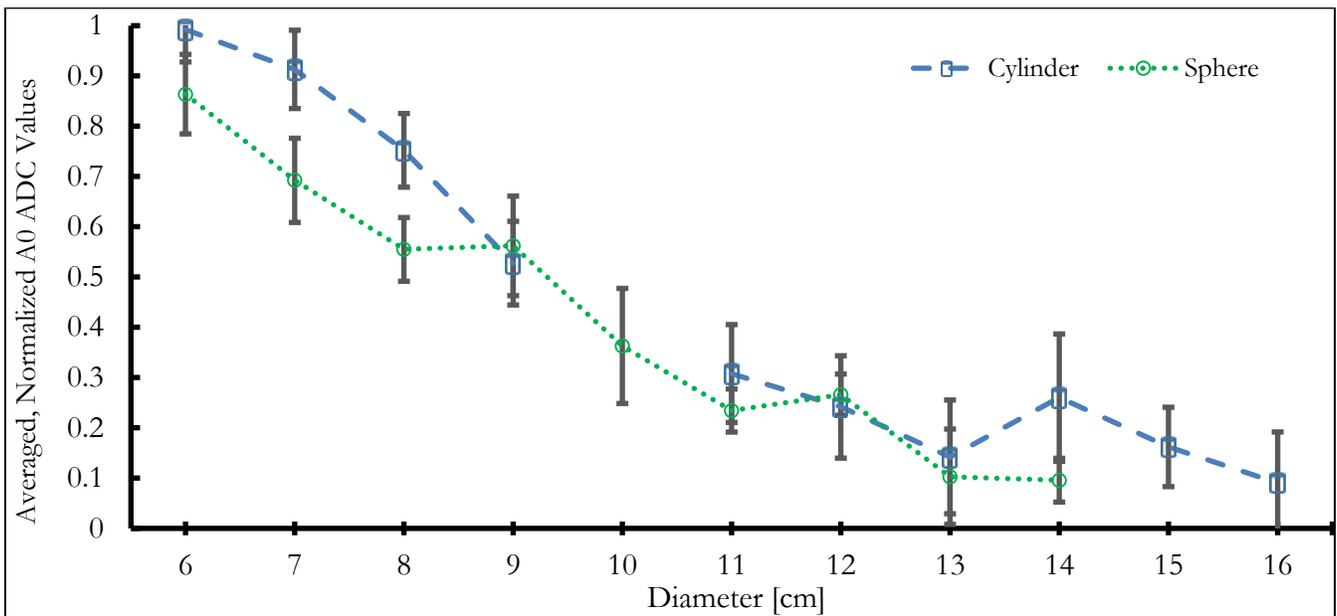

**Figure 9:** Index-Cylinder and Index-Sphere Interaction

Index finger flexion is depicted in **Figure 9** as a function of increasing object diameter. Error bars consist of standard error of the mean. Overlap between error bars occur at 11-13 cm in shape diameter. From 10-16 cm, the Index ADC values did not have a strong linear relationship with cylinder diameter; $R^2$ values for Cylinder diameters greater than 10 cm were less than 0.50.

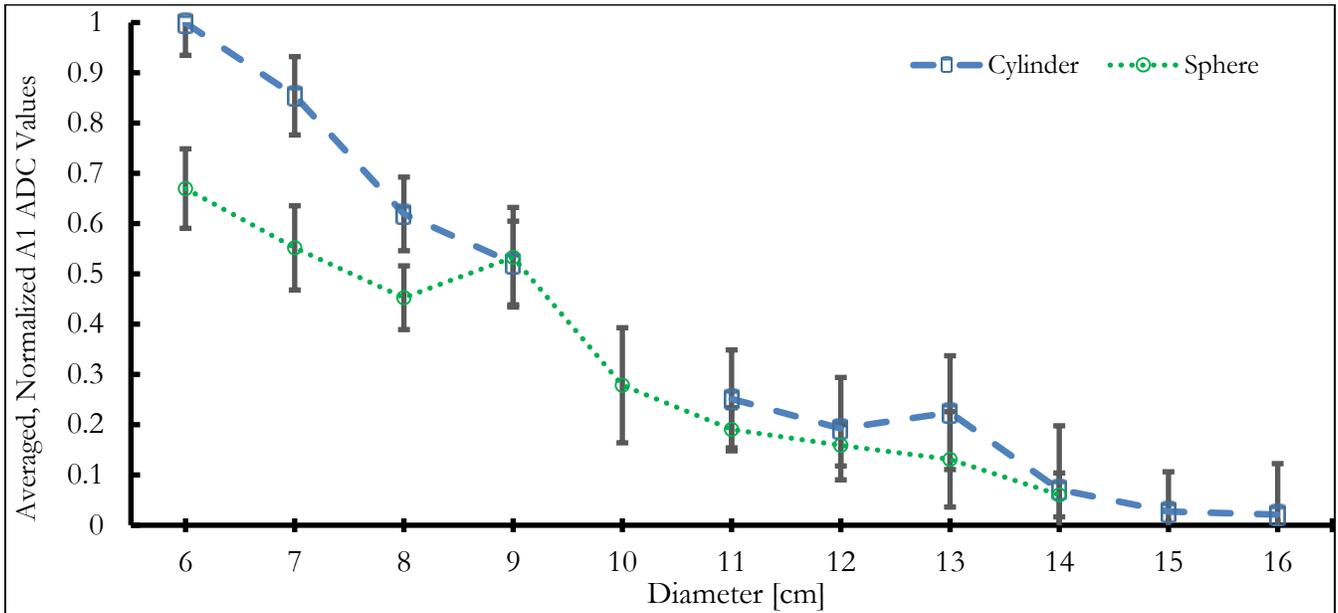

**Figure 10**: Middle-Cylinder and Middle-Sphere Interaction

The plot for **Figure 10** has Middle finger flexion as a function of increasing object diameter. Overlap between error bars occur at 9 and 11-14 cm; the ADC values recorded from 8 cm to 14 cm are very close between Cylinder and Sphere. $R^2$ values were greater than 0.50 for all shapes and their diameters, suggesting a strong linear relationship between object diameter and ADC values for Middle fingers.

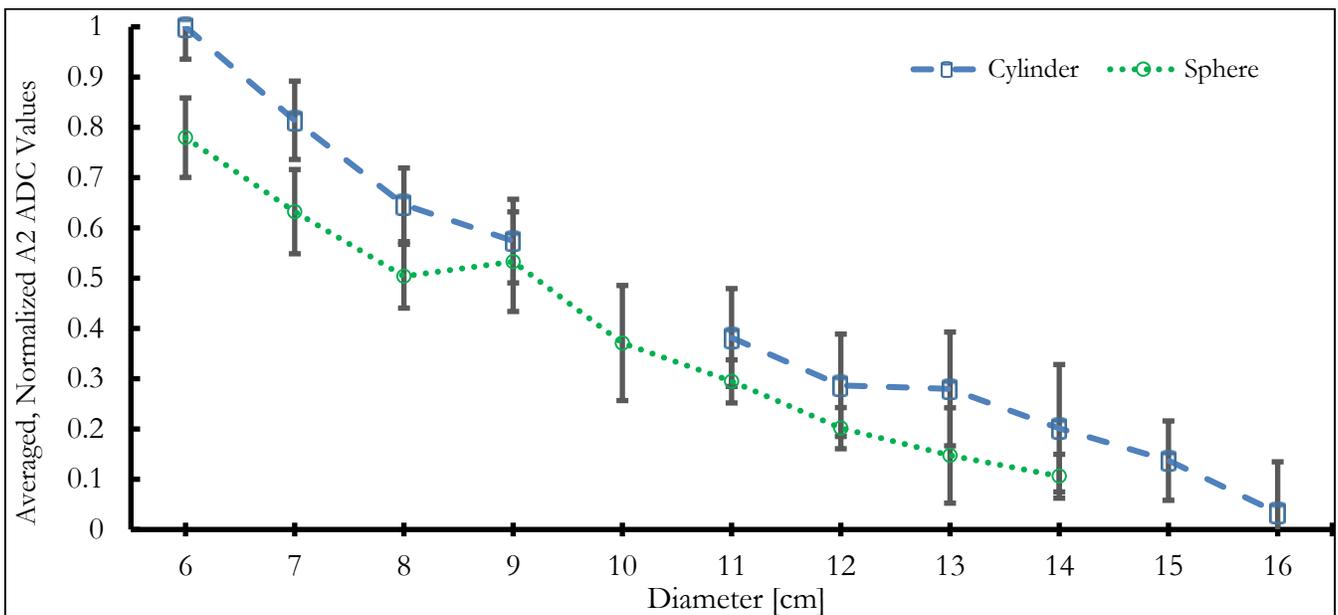

**Figure 11:** Ring-Cylinder and Ring-Sphere Interaction

This plot, **Figure 11**, represents Ring finger flexion over object diameter. Overlap between error bars occur at 9 and 11-14 cm. The Ring finger had the highest $R^2$ values of all fingers, being greater than 0.95 for both shapes; the Ring finger ADC values have the strongest linear correlation with object diameter.

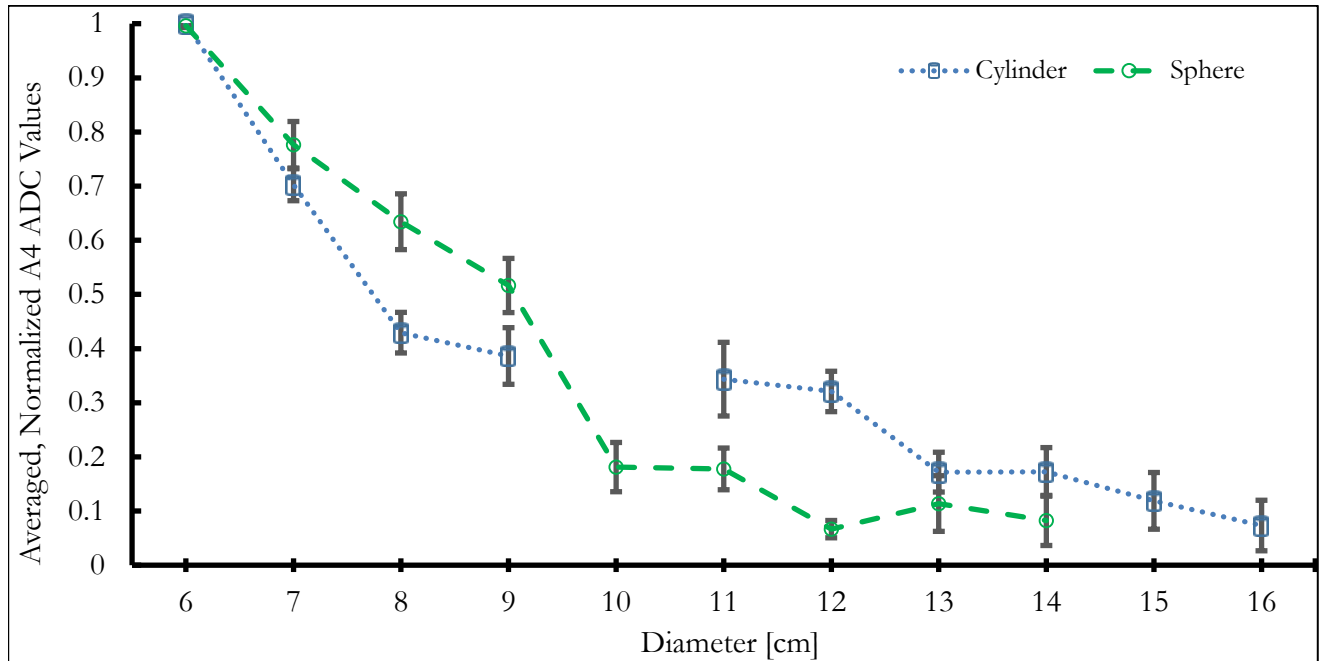

**Figure 12**: Pinky-Cylinder and Pinky-Sphere Interaction

Pinky finger flexion as a function of increasing object diameter. At a diameter of 13 cm, a single error bar overlap occurs. Pinky finger ADC values diverge from the other fingers in that Sphere values can be greater than Cylinder values at diameters less than 10 cm. Of note, the smallest diameters of both shapes have an ADC value of exactly 1.

In **Figures 8-12** the Thumb, Index, Middle, and Ring flex sensors had lower ADC outputs for spherical grasps in comparison to cylindrical grasps. In contrast, the **Figure 12** ADC values collected from the pinky-mounted flex sensor were higher during sphere interaction as opposed to cylinder interaction, unless the sphere was greater than 10 cm diameter. Error bars generated from SEM had overlapping values at 9 cm for all fingers except for the pinky finger. Objects with diameters greater than 10 cm had a higher occurrence of SEM overlap for the Index, Middle, and Ring fingers. Additionally, when objects were larger than 10 cm, recorded ADC values levelled off or did not decrease with an increase in object diameter; $R^2$ values for objects greater than 10 cm were less than 0.50 in **Figures 8 and 9**. However, there is a linear relationship between read ADC values and object size, with a greater than 0.83 $R^2$ value for all finger-object interactions observed.

**Discussion and Conclusion**

Some limitations encountered during this experiment include a lack of collected data at the 10 cm Cylinder diameter, a range mismatch between Sphere and Cylinder diameters, and missing data from a few data glove users due to time limitations caused by the Coronavirus quarantine.

Data glove finger "bend" or ADC values read from the A0-4 Analog Input Pin decrease with an increase in diameter, generally, with the notable exceptions of the Index finger and Thumb at diameters greater than 10 cm (**Figures 8 and 9)**. ADC values read at larger diameters are saturated; the radius of an object must go to infinity for ADC values to get to their "flat" bend angle or minimum ADC values.[9] The Pinky finger in particular (**Figure 12)** has a distinct unique reaction to both shapes; not only do all of its ADC values have the least SEM overlap, but it also bends more in reaction to Cylinder interaction versus its sphere equivalent. Therefore, utilizing all five fingers of a data glove, as opposed to just three,[11] can aid in differentiating objects as seen in **Figures 8-12**.

Having different users incorporated into the data collection process ensured the robustness of the generated data; a large set of users of the same glove system was observed as a common limiting factor in the reliability of grasp classification and data glove performance.[6] Consistent with the literature, there was a correlation between object size and flex sensor readings despite non-linear bending angles when observing objects larger than 10 cm. [8,9] At least one finger had a non-overlapping confidence interval when comparing objects at the same diameter except for 13 cm; when observing the spheres and cylinders of all sizes, all five fingers had a unique reaction to each shape. These results could be used in machine learning or statistical models for real-time object classification.

**References**

1. Tiwana, M. I., Redmond, S. J., & Lovell, N. H. (2012). A review of tactile sensing technologies with applications in biomedical engineering. Sensors and Actuators A: Physical, 179, 17–31. https://doi.org/10.1016/j.sna.2012.02.051
2. Gehler, P., & Nowozin, S. (2009, September). On feature combination for multiclass object classification. 2009 IEEE 12th International Conference on Computer Vision. 2009 IEEE 12th International Conference on Computer Vision (ICCV). https://doi.org/10.1109/iccv.2009.5459169
3. Jones, E. G. (2006). The sensory hand. Brain, 129(12), 3413–3420. https://doi.org/10.1093/brain/awl308
4. Schultz, A. E., Solomon, J. H., Peshkin, M. A., & Hartmann, M. J. (n.d.). Multifunctional Whisker Arrays for Distance Detection, Terrain Mapping, and Object Feature Extraction. Proceedings of the 2005 IEEE International Conference on Robotics and Automation. 2005 IEEE International Conference on Robotics and Automation. https://doi.org/10.1109/robot.2005.1570503
5. Rijpkema, H., & Girard, M. (1991). Computer animation of knowledge-based human grasping. ACM SIGGRAPH Computer Graphics, 25(4), 339–348. https://doi.org/10.1145/127719.122754
6. Heumer, G., Amor, H. B., Weber, M., & Jung, B. (2007). Grasp Recognition with Uncalibrated Data Gloves - A Comparison of Classification Methods. 2007 IEEE Virtual Reality Conference. 2007 IEEE Virtual Reality Conference. https://doi.org/10.1109/vr.2007.352459
7. https://www.spectrasymbol.com/wp-content/uploads/2019/07/flexsensordatasheetv2019revA.pdf
8. Elgeneidy, K., Neumann, G., Pearson, S., Jackson, M., & Lohse, N. (2018, October). Contact Detection and Size Estimation Using a Modular Soft Gripper with Embedded Flex Sensors. 2018 IEEE/RSJ International Conference on Intelligent Robots and Systems (IROS), 498-503. https://doi.org/10.1109/iros.2018.8593399

9. Saggio, G., & Orengo, G. (2018). Flex sensor characterization against shape and curvature changes. Sensors and Actuators A: Physical, 273, 221–231. https://doi.org/10.1016/j.sna.2018.02.035
10. Lin, B.-S., Lee, I.-J., Yang, S.-Y., Lo, Y.-C., Lee, J., & Chen, J.-L. (2018). Design of an Inertial-Sensor-Based Data Glove for Hand Function Evaluation. Sensors, 18(5), 1545. https://doi.org/10.3390/s18051545
11. Nazrul Hamizi, A., Khairunizam, W., Ahmad, W., Shahriman, A. B., Aida, J., & Bakar, A. (2013). Classification of finger grasping by using PCA based on best matching unit (BMU) approach.
12. https://www.arduino.cc/reference/en/language/functions/analog-io/analogread/
13. Fritzing [Computer software]. (2019). Retrieved from https://fritzing.org/
14. Tinkercad [Computer software]. (2020). Retrieved from https://www.tinkercad.com/
15. Arduino Platform. (n.d.). Retrieved October 12, 2020, from https://www.arduino.cc/
16. Han, Jiawei; Kamber, Micheline; Pei, Jian (2011). "Data Transformation and Data Discretization". Data Mining: Concepts and Techniques. Elsevier. pp. 111–118.